\documentclass[10pt,twocolumn,letterpaper]{article}

\usepackage{cvpr}
\usepackage{times}
\usepackage{epsfig}
\usepackage{graphicx}
\usepackage{amsmath}
\usepackage{amssymb}

\usepackage{MnSymbol}%
\usepackage{wasysym}%

\usepackage{latexsym}
\usepackage{url}

\usepackage{multirow}
\usepackage{enumitem}
\usepackage{stfloats}
\usepackage{booktabs}
\usepackage{colortbl}
\usepackage{subcaption}

\usepackage{array}   
\usepackage{graphics}
\usepackage{arydshln}

\usepackage{calc}
\usepackage{comment}

\usepackage{algorithm}
\usepackage{algpseudocode}

\usepackage{amsmath}
\usepackage{amsfonts}
\usepackage{amssymb}
\usepackage{wrapfig}
\usepackage{subcaption}
\usepackage{multirow}
 \usepackage{mathtools} 

\usepackage{verbatim}

\usepackage{anyfontsize}

\usepackage{microtype}
\usepackage{graphicx}
\usepackage{booktabs} 
\usepackage{multirow}
\usepackage{amsmath,amssymb}
\usepackage{booktabs}
\usepackage{caption,subcaption}
\usepackage{xcolor}
\usepackage{enumitem}
\setitemize{itemsep=10pt,topsep=0pt,parsep=0pt,partopsep=0pt}
\usepackage{adjustbox}

\newcommand{\RN}[1]{%
	\textup{\lowercase\expandafter{\it \romannumeral#1}}%
}

\def\Std{\textsf{Std}} 
\def\Beta{\textsf{Beta}} 

\newcommand{\ea}[0]{\emph{et al. }}

\newcommand{\onev}{\ensuremath{{\bf 1}}}

\newcommand{\beq}{\vspace{0mm}\begin{equation}}
\newcommand{\eeq}{\vspace{0mm}\end{equation}}
\newcommand{\beqs}{\vspace{0mm}\begin{eqnarray}}
\newcommand{\eeqs}{\vspace{0mm}\end{eqnarray}}
\newcommand{\barr}{\begin{array}}
\newcommand{\earr}{\end{array}}

\newcommand{\Cmat}{{\bf C}}

\newcommand{\Mmat}{{\bf M}}

\newcommand{\Qmat}[0]{{{\bf Q}}\xspace}

\newcommand{\cv}[0]{{\boldsymbol{c}}}

\newcommand{\hv}[0]{{\boldsymbol{h}}}

\newcommand{\xv}{\boldsymbol{x}}
\newcommand{\yv}{\boldsymbol{y}}
\newcommand{\zv}{\boldsymbol{z}}

\newcommand{\thetav}{\boldsymbol{\theta}}

\newcommand{\R}{\mathbb{R}}
\newcommand{\E}{\mathbb{E}}

\newcommand{\Lcal}{\mathcal{L}}

\newcommand{\Tcal}{\mathcal{T}}

\usepackage{color, colortbl}
\definecolor{Gray}{gray}{0.93}

\newcommand{\shortname}{\textsc{Hexa}}

\usepackage{xcolor}
\usepackage{color, colortbl}
\newcommand{\blackcircle}{
~{\leavevmode\put(0,2.83){\circle*{5.5}}}~
}

\usepackage[pagebackref=true,breaklinks=true,letterpaper=true,colorlinks,bookmarks=false]{hyperref}

\cvprfinalcopy 

\def\cvprPaperID{8841} 

\ifcvprfinal\fi
\begin{document}

\title{Self-supervised Pre-training with Hard Examples\\ Improves Visual Representations}

\author{
Chunyuan Li$^{1}$,  ~~Xiujun Li$^{1}$,  ~~Lei Zhang$^{1}$,  ~~Baolin Peng$^{1}$, ~~Mingyuan Zhou$^{2}$,  ~~Jianfeng Gao$^{1}$ \\  $^{1}$Microsoft Research, Redmond ~~~  $^{2}$The University of Texas at Austin\\
{\tt\small   \{chunyl,xiul,leizhang,bapeng,jfgao\}@microsoft.com~ \{mingyuan.zhou\}@mccombs.utexas.edu}
}

\maketitle

\begin{abstract}


Self-supervised pre-training (SSP) employs random image transformations to generate training data for visual representation learning. In this paper, we first present a modeling framework that unifies existing SSP methods as learning to predict pseudo-labels. Then, we propose new data augmentation methods of generating training examples whose pseudo-labels are harder to predict than those generated via random image transformations. Specifically, we use adversarial training and CutMix to create hard examples (HEXA) to be used as augmented views for MoCo-v2 and DeepCluster-v2, leading to two variants \shortname{}$_{\text{MoCo}}$ and \shortname{}$_{\text{DCluster}}$, respectively. In our experiments, we pre-train models on ImageNet and evaluate them on multiple public benchmarks. Our evaluation shows that the two new algorithm variants outperform their original counterparts, and achieve new state-of-the-art on a wide range of tasks where limited task supervision is available for fine-tuning. These results verify that hard examples are instrumental in improving the generalization of the pre-trained models.    

\end{abstract}

\section{Introduction}

Self-supervised visual representation learning aims to learn image features from raw pixels without relying on 
manual supervisions.
Recent results show that self-supervised pre-training (SSP) outperforms state-of-the-art (SoTA) fully-supervised pre-training methods~\cite{he2020momentum,caron2020unsupervised}, and is becoming
the building block in many computer vision applications.
The pre-trained model produces general-purpose features and serve as the backbone of 
various downstream tasks such as classification, detection and segmentation, improving the generalization of those task-specific models that are often trained on limited amounts of task labels. 

Most state-of-the-art SSP methods focus on designing novel pretext objectives, ranging from the traditional prototype learning~\cite{xie2016unsupervised,yang2016joint,caron2018deepcluster,ji2019invariant,zhan2020online}, to a recently popular concept known as contrastive learning~\cite{chen2020simple,he2020momentum,chen2020improved,grill2020bootstrap}, and a combination of both~\cite{caron2020unsupervised,li2020prototypical}. Apart from the improved efficiency, all these methods heavily rely on data augmentation to create different views of an image using image transformations, such as random crop (with flip and resize), color distortion, and Gaussian blur. 
Recent studies show that SSP performance can be further improved by using more aggressively transformed views, such as increasing the number of views~\cite{caron2020unsupervised}, and more distinctive views via minimizing mutual information~\cite{tian2020makes}.

However, image transformations are agnostic to the pretext objectives, and it remains unknown how to augment views specifically based on the 
pre-training tasks themselves, 
and how different augmentation methods affect the generalization of the learned models.
To tailor data augmentation to pre-training tasks,
we explicitly formulate SSP as a problem of predicting pseudo-labels, based on which we propose to generate {\em hard examples} (\shortname{}), 
a family of augmented views whose pseudo-labels are difficult to predict. 
Specifically, two schemes are considered. 
$(\RN{1})$ Adversarial examples are created with the intention to cause an SSP model to make prediction mistakes and thus improve the generalization of the model~\cite{xie2020adversarial}.
$(\RN{2})$ Cut-mixed examples are created via cutting and pasting patches among different images~\cite{yun2019cutmix}, so that its content is a mixture of multiple images.

Our contributions include:
$(\RN{1})$ A pseudo-label perspective is formulated to motivate the concept of hard examples in self-supervised learning.
$(\RN{2})$ Two novel algorithms are proposed, through applying our framework to two distinctly different existing approaches.
$(\RN{3})$ Experiment are conducted on a wide range of tasks in self-supervised benchmarks, showing that  \shortname{}  consistently improves their original counterparts, and achieves SoTA performance under the same settings. It demonstrates the genericity and effectiveness of proposed framework in constructing hard examples for improving the visual representations using SSP.

\section{Self-supervision: A Pseudo-label View}

Self-supervised learning learns representations
by leveraging the weak signals intrinsically existing in images as {\em pseudo-labels}, and  
maximizing agreement between pseudo-labels and the learned representations. This framework
comprises the following four major components~\cite{chen2020simple}.
$(\RN{1})$ {\em Data augmentation} that randomly transforms
any given image $\xv$, resulting in multiple correlated views of the same example, denoted as $\{ \Tilde{\xv}_i \}$.
$(\RN{2})$ A {\em backbone network} $\hv = f_{\thetav_0}(\Tilde{\xv})$ parameterized by $\thetav_0$ that extracts a feature representation $\hv \in \R^d$ from an augmented view $ \Tilde{\xv}$.
$(\RN{3})$ A {\em projection head} $\zv = f_{\thetav_1}(\hv)$ parameterized by $\thetav_1$ that maps the feature
representation $\hv$ to a latent representation $\zv$, on which self-supervised loss is
applied. 
$(\RN{4})$
A {\em  self-supervised loss} function aims to predict the {\em pseudo-label} $\yv$ based on $\zv$. Different self-supervised learning methods differ in their exploited weak signals, based on which different kinds of pseudo-labels are constructed. We cast a broad family of SSP methods as a pseudo-label classification task, where $\{\thetav_0,\thetav_1\}$ are the classifier parameters; After training, $\thetav_0$ provides generic visual representations.
Following this point of view, we revisit two types of methods. 

\paragraph{Type I: Contrastive Learning.}
Contrastive learning is a framework that learns representations
by maximizing agreement between differently augmented
views of the same image via a contrastive loss in
the latent space.
For a given {\em query} $\zv_q$,  we identify its positive samples $\zv_{k^+}$ from a set of {\em keys} $\{\zv_k\} = \{\zv_{k^+}, \zv_{k^-} \}$, where positive samples are indexed by $k^+$ and negative samples are indexed by $k^-$. The pseudo-labels in contrastive learning are defined by feature pairwise comparisons: $y=1$ for the pair $(\zv_q, \zv_{k^+})$ and $y=0$ for the pair $(\zv_q, \zv_{k^-})$. For a query with $K$ pairs, its pseudo-label vector is $\yv \in \{0,1\}^K$.

The contrastive prediction task can be formulated as a dictionary look-up problem. By mapping a view $\Tilde{\xv}$ into a latent representation $\zv$ using the function composition $\zv = f_{\thetav} (\Tilde{\xv})= f_{\thetav_1} \circ f_{\thetav_0} (\Tilde{\xv})$,
an effective contrastive loss function, called InfoNCE, can be derived as:
\begin{align} 
\hspace{-0mm}
& \min_{\thetav} \Lcal^{\text{Std}}_{\text{C}} ( \Tilde{\xv}_q, \Tilde{\xv}_{k}, \yv) \nonumber\\
=& \min_{\thetav} - \sum_{k} y_k \log \frac{\exp(f_{\thetav} (\Tilde{\xv}_q) \! \cdot\! f_{\thetav} (\Tilde{\xv}_{ k})/\tau)}{ \sum_{k'}\exp(f_{\thetav} (\Tilde{\xv}_q) \!\cdot\! f_{\thetav} (\Tilde{\xv}_{k'}) /\tau) }  
\label{eq_info_nce1} \\
=& \min_{\thetav}  -\log \frac{\exp(f_{\thetav} (\Tilde{\xv}_q) \! \cdot\! f_{\thetav} (\Tilde{\xv}_{ k^+})/\tau)}{ \sum_{k'}\exp(f_{\thetav} (\Tilde{\xv}_q) \!\cdot\! f_{\thetav} (\Tilde{\xv}_{k'}) /\tau) }
\label{eq_info_nce2}
\end{align}
where $\thetav = \{\thetav_0, \thetav_1\}$ denotes the set of  trainable parameters  and $\tau$ is a temperature hyper-parameter. From~\eqref{eq_info_nce1} to~\eqref{eq_info_nce2}, only the loss term indexed with $ k^+$ remains, while the ones indexed with $k^-$ are excluded, because their corresponding pseudo-label $y=0$.

In the instance discrimination pretext task (used by MoCo and SimCLR), a query and a key form a positive pair if they are data-augmented versions of the same image, and otherwise form a negative pair.
The contrastive loss~\eqref{eq_info_nce2} can be minimized by various
mechanisms that differ in how the keys (or negative samples) are maintained~\cite{chen2020improved}.

\begin{itemize}
\vspace{1mm}
    \item {\bf SimCLR}~\cite{chen2020simple}~ The negative keys are from the same batch and updated end-to-end by back-propagation. SimCLR is based on this
mechanism and requires a large batch to provide a large set
of negatives. 
    \vspace{-0mm}
    \item  {\bf MoCo}~\cite{he2020momentum,chen2020improved}~
In the MoCo mechanism, the negative keys are maintained in a queue $\Qmat$, and only the queries and positive keys are encoded in each training batch. A momentum encoder is adopted to improve the representation consistency between the current and earlier keys. MoCo decouples the batch size from the number of negatives. MoCo-v2~\cite{chen2020improved} is an improved version using strong augmentation (\ie more aggressive image transformations) and MLP projection proposed in SimCLR.
    \vspace{-6mm}
\end{itemize}

\paragraph{Type II: Prototype Learning.}
The prototype learning methods~\cite{caron2018deepcluster,li2020prototypical,caron2020unsupervised} introduce a ``prototype'' as the centroid for a cluster formed by similar image views. 
The latent representations are fed into a clustering algorithm to produce the prototype/cluster assignments, which are subsequently used as ``pseudo-labels'' to supervise representation learning. 

{\bf  DeepCluster} is a representative prototype learning work. It employs $K$-means as the clustering algorithm, which takes a set of latent vectors $\zv = f_{\thetav} (\xv)$ as input, clusters them into $K$ distinct groups with prototypes $\Cmat$,  and simultaneously output the optimal cluster assignment $ \yv  \in \Delta^K$ as a one-hot probability simplex. The model is trained to predict the optimal assignment:
\begin{align} 
\hspace{-0mm}
& \min_{\thetav} \Lcal^{\text{Std}}_{\text{P}} ( \Tilde{\xv}, \yv) \nonumber\\
=& \min_{\thetav} - \sum_{j=1}^K y_{j} \log \frac{\exp(f_{\thetav} (\Tilde{\xv}_j)  \cdot \cv_{j} \! /\tau)}{ \sum_{j'}\exp(f_{\thetav} (\Tilde{\xv}_{j'}) \cdot \cv_{j'} /\tau)  }  
\label{eq_deep_cluster1} \\
=& \min_{\thetav}  -\log \frac{\exp(f_{\thetav} (\Tilde{\xv}_{j^*})  \cdot \cv_{j^*} \! /\tau)}{ \sum_{j'}\exp(f_{\thetav} (\Tilde{\xv}_{j'})  \cdot \cv_{j'} /\tau) }  
\label{eq_deep_cluster2}
\end{align}
%
where $\cv_j$ is $j$th prototype/cluster centroid, and the $j^*$ is the index of the assigned cluster for $\Tilde{\xv}$. DeepCluster alternates between two steps: feature clustering using $K$-means and feature learning by predicting these pseudo-labels.
The prototype learning improves limitations of the contrastive instance discrimination methods via allowing views with similar semantics but from different source images to be pushed together.

Note that \eqref{eq_info_nce2} and \eqref{eq_deep_cluster2} represent a traditional view for self-supervised learning formulations, while \eqref{eq_info_nce1} and \eqref{eq_deep_cluster1} are our derived pseudo-label view, where $\yv$ is explicitly involved in the learning objectives. It opens opportunities to study new data augmentation $\Tilde{\xv}$ based on $\yv$, in improving the robustness and generalization of $\thetav_0$ for visual representations. 

\begin{figure*}[t!]
	\vspace{-0mm}\centering
	\begin{tabular}{c c}
		\hspace{-4mm}
		\includegraphics[height=4.3cm]{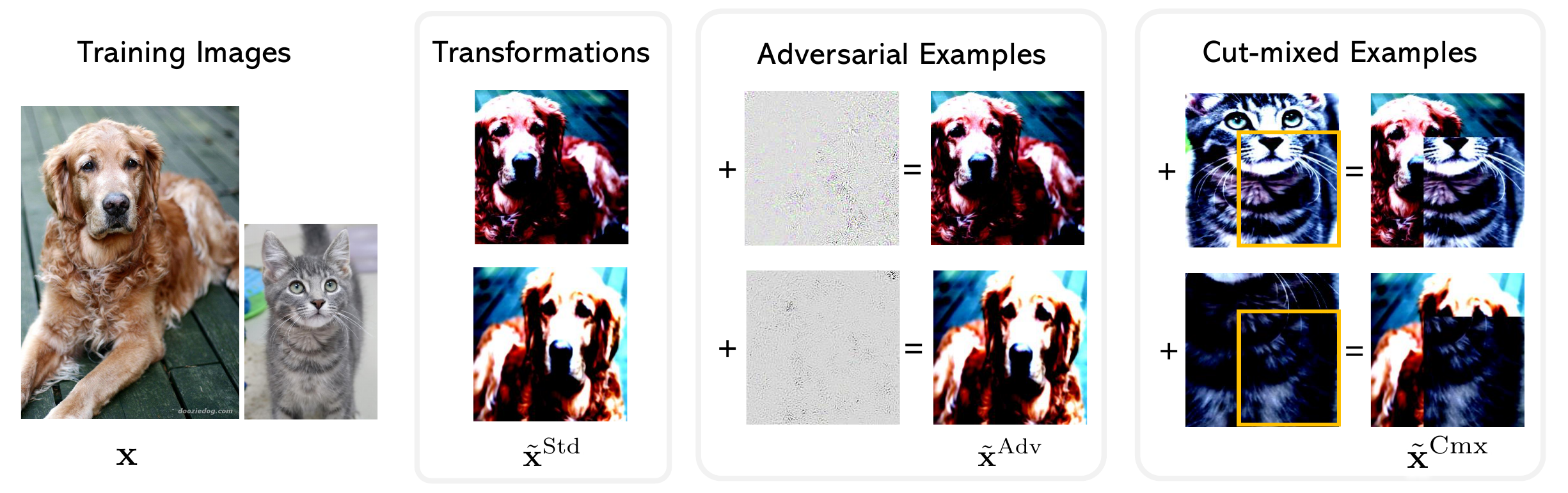}  & 
		\hspace{-6mm}
		\includegraphics[height=4.0cm]{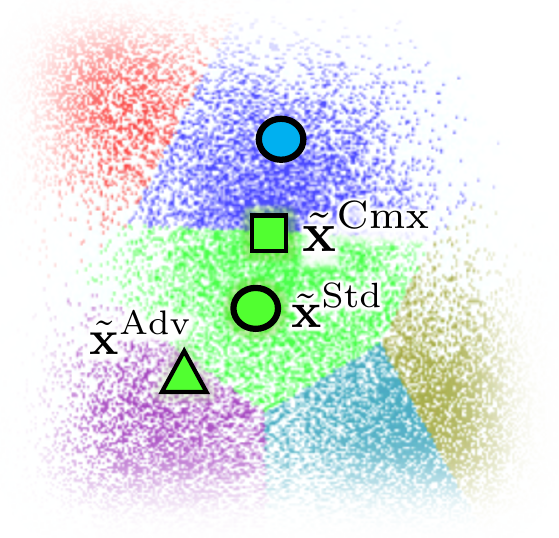} \\
		(a) Image transformations {\em v.s.} Hard examples \vspace{2mm} & 
		\hspace{-6mm}
		(b) Augmented view space \hspace{-0mm} \\ 
	\end{tabular}
	\vspace{-3mm}
	\caption{Illustration of \shortname{}: (a) Hard examples. For the original dog image, existing SSP methods employ random transformations to generate augmented example $\Tilde{\xv}^{\text{Std}}$, we propose two types of hard examples. Adversarial examples $\Tilde{\xv}^{\text{Adv}}$ add perturbations on $\Tilde{\xv}^{\text{Std}}$ and cut-mixed examples $\Tilde{\xv}^{\text{Cmx}}$ cut and paste patches between $\Tilde{\xv}^{\text{Std}}$. (b) A visualization example of augmented view space. Each circle ``$\blackcircle{}$'' indicates an augmented example $\Tilde{\xv}^{\text{Std}}$. The adversarial example $\Tilde{\xv}^{\text{Adv}}$ (``$\blacktriangle$'') fools the model to make a prediction mistake, and the cut-mixed example $\Tilde{\xv}^{\text{Cmx}}$ (``$\blacksquare$'') is created between two standard augmentations.
	 }
	\vspace{-3mm}
	\label{fig:hard_example_illustration}
\end{figure*}

\section{Pre-training with Hard Examples}
Augmented views $\Tilde{\xv}$ play a vital role in SSP. Most existing methods synthesize views through random image transformations, without carefully considering their feasibility in completing the self-supervised learning task: predicting pseudo-labels. 
By contrast, we focus on studying {\em hard examples}, which are defined as augmented views $ \Tilde{\xv}$ whose pseudo-labels $\yv$ are difficult to be predicted. Specifically, we consider two schemes: adversarial examples and cut-mixed examples. We visually illustrate in Figure~\ref{fig:hard_example_illustration} how hard examples are constructed from image transformations, detail the derivation process as follows. 

\subsection{Adversarial Examples}
Adversarial robustness refers to a model’s invariance to small (often imperceptible) perturbations of its inputs (\ie clean examples). The {\em adversarial examples} are produced by adding perturbations on clean examples to fool the predictions of a trained model the most. In self-supervised learning,  we propose to add perturbations on the augmented views $\Tilde{\xv}$ to fool their predicted pseudo-labels. 


\paragraph{Adversarial Contrastive Learning.}
For the instance contrastive discrimination methods, we focus on the MoCo algorithm. Specifically, we propose to generate adversarial examples for query $\Tilde{\xv}_q$ only. Since both key $\Tilde{\xv}_{k}$ and pseudo-label $y$ are fixed, it is feasible to compute the gradient on the query $\Tilde{\xv}_q$, leading to the adversarial training objective:  
\begin{align} 
\hspace{-0mm}
& \min_{\thetav} \Lcal^{\text{Adv}}_{\text{C}} ( \Tilde{\xv}_q, \Tilde{\xv}_{k}, \yv) = \nonumber\\
& \min_{\thetav} \max_{ \| \delta \|_{2} \le \epsilon } 
-\log \frac{\exp(f_{\thetav} (\Tilde{\xv}_q + \delta) \! \cdot\! f_{\thetav} (\Tilde{\xv}_{ k^+})/\tau)}{ \sum_{k'}\exp(f_{\thetav} (\Tilde{\xv}_q + \delta) \!\cdot\! f_{\thetav} (\Tilde{\xv}_{k'}) /\tau) }
\label{eq_info_nce_sat}
\end{align}
where $\epsilon$ is a hyper-parameter governing how invariant the resulting model should be to adversarial attacks, 
and $\delta$ is the perturbation. In practice, \eqref{eq_info_nce_sat} is updated using two steps:  
$(\RN{1})$ 
By applying Projected Gradient Descent (PGD)~\cite{goodfellow2014explaining,bubeck2014convex}, we obtain adversarial examples on-the-fly:
\begin{align} 
\hspace{-0mm}
\Tilde{\xv}^{\text{Adv}}_q  = \Tilde{\xv}_q + \delta 
 = \Tilde{\xv}_q + \eta ~ \text{sign} ( \nabla_{\Tilde{\xv}_q} \Lcal^{\text{Std}}_{\text{C}} ( \Tilde{\xv}_q, \Tilde{\xv}_{k}, \yv) ), 
\label{eq_info_nce_adv_example}
\end{align}
where $ \eta $ the step size for PGD.
The generated $\Tilde{\xv}^{\text{Adv}}_q$ can be viewed as a new data augmentation on query $\Tilde{\xv}_q$, and $(\RN{2})$  we then feed  $\Tilde{\xv}^{\text{Adv}}_q$ into the model to update parameters $\thetav$. Note $\Tilde{\xv}^{\text{Adv}}_q$ differs from traditional random augmentations in that it  takes into consideration of the relationship between the positive example $\Tilde{\xv}_{k}$ and all negative examples within the memory bank, and tends to be a ``harder'' query than $\Tilde{\xv}_q$  for the dictionary look-up problem. 
The adversarial examples for SimCLR is easier to construct, and can be viewed as a special case when all negative examples are from current batch, rather than from memory bank.

\paragraph{Adversarial Prototype Learning.}
The adversarial training for prototype-based methods are similar to supervised settings, after the cluster assignments $\yv$ are learned. We treat these pseudo-labels as targets to fool the model:\vspace{-2mm}
\begin{align} 
\vspace{-4mm}
& \min_{\thetav} \Lcal^{\text{Adv}}_{\text{P}} ( \Tilde{\xv}, \yv)= \nonumber\\
\vspace{-1mm}
& \min_{\thetav}  \max_{ \| \delta \|_{2} \le \epsilon }  -\log \frac{\exp(f_{\thetav} (\Tilde{\xv}_{j^*} + \delta)  \cdot \cv_{j^*}  \! /\tau)}{ \sum_{j'}\exp(f_{\thetav} (\Tilde{\xv}_{j'} + \delta) \cdot \cv_{j'}   ) /\tau) }  
\vspace{-1mm}
\label{eq_deep_cluster_sat}
\end{align}
%
Similarly, ~\eqref{eq_deep_cluster_sat} is also updated in two steps: starting with adversarial example generation, followed by model update. The adversarial example $\Tilde{\xv}^{\text{Adv}}_{j^*} = \Tilde{\xv}_{j^*} + \delta $ is ``harder'' than $\Tilde{\xv}_{j^*}$ to be correctly aligned into clusters. 

\paragraph{Implementations.} It is shown in AdvProp~\cite{xie2020adversarial} that clean examples and adversarial examples tend to have different batch statistics, due to their salient empirical distribution divergence. Thus, we adopt the AdvProp training scheme, where two separate sets of batch normalization (BN)~\cite{ioffe2015batch} parameters are considered, 
summarizing the statistic for clean examples and that for adversarial examples, respectively. In Figure~\ref{fig:hard_example_illustration}, we use the dog image as input and visualize the perturbations 
as noisy grey maps, which are added on $\Tilde{\xv}^{\text{Std}}$. Though $\Tilde{\xv}^{\text{Adv}}$ look indistinguishable with $\Tilde{\xv}^{\text{Std}}$ visually, their corresponding pseudo-labels have been revised significantly, depending on how much they move across the decision boundary (\ie how many PGD steps are applied). We study the impact of hyper-parameters in PGD in Appendix, and we choose PGD step as $1$ for computational efficiency, perturbation threshold $\epsilon=1$ and step size $ \eta =1$.  


\begin{algorithm}[!t]
   \caption{\shortname{}$_{\text{MoCo}}$ }
   \label{alg:sat_moco}
\begin{algorithmic}[1]
    \Require Initializing network parameters for query $\thetav$ and key $\thetav'$; Random image transformations $\mathcal{T}$ and $\mathcal{T}'$; A queue $\Qmat$ for memory bank, with momentum decay coefficient $\beta=0.99$.
   \For{a number of training iterations}
   
        \State {\quad {\small \color{blue} $\#  \mathtt{~Produce~a~mini batch~ of~query~and~key~samples~}$}}    
        \State {\quad Sample a batch of image $\xv$ from the full dataset};
        \State {\quad Clean query $\Tilde{\xv}_q = \Tcal(\xv)$ and key $\Tilde{\xv}_k = \Tcal'(\xv)$};
        
        \State{\quad {\small \color{blue} $\# \mathtt{~Adversarial~example~generation}$}}  
        \State {\quad Forward  $\Tilde{\zv}_q = f_{\thetav}(\Tilde{\xv}_q)$ and $\Tilde{\zv}_k = f_{\thetav'}(\Tilde{\xv}_k)$};
        \State{\quad Synthesize adversarial query $\Tilde{\xv}^{\text{Adv}}_q$ using~\eqref{eq_info_nce_adv_example}};

        \State{\quad {\small \color{blue} $\# \mathtt{~Cutmixed~example~generation}$}}  
        \State{\quad Synthesize ($\Tilde{\xv}^{\text{Cmx}}$, $\yv^{\text{Cmx}}$) using~\eqref{eq_cutmix_example}};

        \State{\quad {\small \color{blue} $\# \mathtt{~Update~query~network}$}} 
        \State {\quad Forward  $\Tilde{\xv}_q$, $\Tilde{\xv}^{\text{Adv}}_q$ and $\Tilde{\xv}^{\text{Cmx}}$ using~\eqref{eq_sat_overall_loss}};
        \State {\quad Compute $\frac{\partial \Lcal^{\rm \shortname{}}_{\rm C} }{\partial \thetav }$ of~\eqref{eq_sat_overall_loss} and update $\thetav$ };
        
        \State{\quad {\small \color{blue} $\# \mathtt{~Update~key~network~with~momentum}$}} 
        \State {\quad $\thetav' \leftarrow \beta \thetav' + (1-\beta) \thetav$};

        \State{\quad {\small \color{blue} $\# \mathtt{~Update~memory~bank}$}} 
        \State {\quad Queuing $\Tilde{\zv}_k$ in $\Qmat$ and dequeuing oldest elements};
   \EndFor
\end{algorithmic}
\end{algorithm}

\vspace{-0mm}
\subsection{Cut-Mixed Examples}

Cutmix~\cite{yun2019cutmix} is a recent image augmentation technique for supervised learning. Patches are cut and pasted among images to create a new example, where the ground truth labels are also mixed proportionally to the area of the patches.
Specifically, for randomly selected two images, we consider an augmented view from  $(\Tilde{\xv}_a, \yv_a)$ and $(\Tilde{\xv}_b, \yv_b)$, where $\yv_a$ and $\yv_b$ are the corresponding pseudo-labels (\ie instance identity in contrast learning or cluster index in prototype learning). 
The {\em cutmixed example} $\Tilde{\xv}^{\text{Cmx}}$ and its pseudo-labels $\yv^{\text{Cmx}}$ are generated using the combining operation as:
\begin{align} 
\hspace{-0mm}
\Tilde{\xv}^{\text{Cmx}}&  = \Mmat \odot  \Tilde{\xv}_a +  (\onev-\Mmat) \odot  \Tilde{\xv}_b \nonumber \\
\yv^{\text{Cmx}} & = \lambda \yv_a + (1- \lambda ) \yv_b
\label{eq_cutmix_example}
\end{align}
where $\Mmat\!\in\!\{0,1\}^{W\times H}$ (width $W$ and height $H$) denotes a binary mask indicating where to drop out and fill in from two images, $ \onev $ is a binary
mask filled with ones, $\odot$ is element-wise multiplication, $\lambda$ is the combination ratio between two views. Following~\cite{yun2019cutmix}, $\lambda$ is initially sampled from the beta distribution $\Beta(\alpha, \beta)$, and is finally set as the area percentage that view $\Tilde{\xv}_a$ occupies in $\Tilde{\xv}^{\text{Cmx}}$. The beta distribution controls how much the two views are mixed. We empirically study  hyper-parameters of $(\alpha, \beta)$ in Appendix, and use $\Beta(5, 3)$ in our experiments, which leads to the mean $\E[\lambda]=0.62$ and standard deviation $\Std[\lambda]=0.16$.

Since $\Tilde{\xv}^{\text{Cmx}}$ has mixed contents from two different source images, it tends to be a hard example in predicting either of its labels. The added patches further enhance the localization ability by requiring the
model to identify the object from a partial view.
To train the model, an objective can be written with the standard loss function as:
\begin{align} 
\hspace{-0mm}
& \min_{\thetav} \Lcal^{\text{Cmx}} ( \Tilde{\xv}^{\text{Cmx}} , \yv^{\text{Cmx}}) = \nonumber \\
& \min_{\thetav}  \lambda \Lcal^{\text{Std}} ( \Tilde{\xv}^{\text{Cmx}}, \yv_a) +   (1- \lambda ) \Lcal^{\text{Std}} (\Tilde{\xv}^{\text{Cmx}}, \yv_b)
\label{eq_cutmix_obj}
\end{align}
%

\begin{algorithm}[!t]
   \caption{\shortname{}$_{\text{DCluster}}$ }
   \label{alg:sat_deepcluster}
\begin{algorithmic}[1]
    \Require Initializing network parameters $\thetav$; Random image transformations $\mathcal{T}$; Initializing a set of prototypes $\Cmat$ and compute initial assignment $\yv$.
   \For{a number of training epoch}

   \For{a number of training iteration}   
        \State {\quad {\small \color{blue} $\#  \mathtt{~Produce~a~mini batch~of~samples~}$}}    
        \State {\quad Sample image batch $\xv$ from the full dataset};
        \State {\quad Augmented views $\Tilde{\xv} = \Tcal(\xv)$ };
        
        \State{\quad {\small \color{blue} $\# \mathtt{~Adversarial~example~generation}$}}  
        \State {\quad Forward  $\Tilde{\zv} = f_{\thetav}(\Tilde{\xv})$};
        \State{\quad Synthesize $\Tilde{\xv}^{\text{Adv}}_{j^*}$ using~\eqref{eq_deep_cluster_sat}};

        \State{\quad {\small \color{blue} $\# \mathtt{~Cutmixed~example~generation}$}}  
        \State{\quad Synthesize ($\Tilde{\xv}^{\text{Cmx}}$, $\yv^{\text{Cmx}}$) using~\eqref{eq_cutmix_example}};

        \State{\quad {\small \color{blue} $\# \mathtt{~Update~network}$}} 
        \State {\quad Forward  $\Tilde{\xv}$, $\Tilde{\xv}^{\text{Adv}}_{j^*}$ and $\Tilde{\xv}^{\text{Cmx}}$ using~\eqref{eq_sat_overall_loss}};
        \State {\quad Compute $\frac{\partial \Lcal^{\rm \shortname{}}_{\rm P} }{\partial \thetav }$ of~\eqref{eq_sat_overall_loss} and update $\thetav$};
   \EndFor        
        \State{\!\! {\small \color{blue} $\# \mathtt{~Update~assignment/ pseudolabel~for~each~image}$}} 
        \State{ \!\! Collect $\Tilde{\zv}$ in above inner-loop};          
        \State{ \!\! Solve $K$-means to update $\Cmat$ and $\yv$ for each $\xv$};    
   \EndFor
\end{algorithmic}
\end{algorithm}
\vspace{-3mm}

\paragraph{Implementations.} In each training iteration, we consider images in the original batch as $\Tilde{\xv}_a$,  randomly permute images in a batch to create $\Tilde{\xv}_b$, and generate 
cut-mixed samples 
by combining selected examples from two batches with the same index, according to~\eqref{eq_cutmix_example}. For MoCo, we perform cutmix on queries, and leave keys unchanged. When multiple crops are considered for each image in DeepCluster, the same permutation index is shared among the crops.  In Figure~\ref{fig:hard_example_illustration}, transformations on dog image are $\Tilde{\xv}_a$ ,  transformations on cat image are $\Tilde{\xv}_b$. The cut-mixed examples $\Tilde{\xv}^{\text{Cmx}}$ are ``dog-cat'' images shown on the right side of Figure~\ref{fig:hard_example_illustration}(a). One may imagine $\Tilde{\xv}^{\text{Cmx}}$ often lies at the decision boundary, depending on how much content is mixed from each.

\subsection{Full \shortname{} Objective}
The overall self-adversarial training objective considers both clean and hard examples constructed by adversarial and cutmix augmentations:
\begin{align} 
\hspace{-0mm}
\min_{\thetav}  \Lcal^{\text{\shortname{}}} =  \Lcal^{\text{Std}}  +  \alpha_1 \Lcal^{\text{Adv}} +  \alpha_2 \Lcal^{\text{Cmx}} 
\label{eq_sat_overall_loss}
\end{align}
where $\alpha_1$ and $\alpha_2$ are the weighting hyper-parameters to control the effect of adversarial examples and cutmixed examples, respectively. In our experiments, we set $\alpha_1\!\!  =\!1$ and/or $\!\alpha_2\!\! =\!1$. Note that $\alpha_1\!\! =\! \alpha_2\! \!=\!0$ reduces the objective to the standard self-supervised training algorithms. Concretely, we consider two novel algorithms: 

\begin{itemize}
    \item {\bf \shortname{}$_{\text{MoCo}}$}~By plugging terms \eqref{eq_info_nce2} \eqref{eq_cutmix_example} and \eqref{eq_info_nce_sat} into \eqref{eq_sat_overall_loss}, it yields the full self-adversarial   contrastive learning objective denoted as $\Lcal^{\rm \shortname{}}_{\rm C}$. The \shortname{}$_{\text{MoCo}}$ training procedure is detailed in Algorithm~\ref{alg:sat_moco}. We build \shortname{}$_{\text{MoCo}}$ on top of MoCo-v2. The two algorithms are distinguished 
    from each other in Lines~5-9, where hard examples are computed on query and subsequently employed in model update for \shortname{}$_{\text{MoCo}}$.
    \vspace{-0mm}
    \item  {\bf \shortname{}$_{\text{DCluster}}$}~The  full  self-adversarial prototype learning objective $\Lcal^{\rm \shortname{}}_{\rm P}$ is obtained via plugging \eqref{eq_deep_cluster2}\eqref{eq_cutmix_example} and \eqref{eq_deep_cluster_sat} into \eqref{eq_sat_overall_loss}.  We build \shortname{}$_{\text{DCluster}}$ based on DeepCluster-v2~\cite{li2020prototypical}, which improves DeepCluster~\cite{caron2018deepcluster} to reach similar performance with recent state-of-the-art methods. The \shortname{}$_{\text{DeepCluster}}$ training procedure is detailed in Algorithm~\ref{alg:sat_deepcluster}. It differs from DeepCluster-v2 in Lines 6-10, where hard examples are computed to train the network in conjunction with clean examples.
    \vspace{-0mm}
\end{itemize}
\textbf{}

\vspace{-4mm}
\section{Related Works}
\vspace{-1mm}
\subsection{Self-supervised Pre-training} 
\paragraph{Pretext task taxonomy.}
Self-supervised learning is a popular form of unsupervised learning, where  labels annotated by humans are replaced by ``pseudo-labels'' directly extracted from the raw input data by leveraging its intrinsic structures. We broadly categorize existing self-supervised learning methods into three classes: 
$(\RN{1})$
{\em Handcrafted pretext tasks}. This includes many traditional self-supervised methods such as relative position of patches~\cite{doersch2015unsupervised,noroozi2016unsupervised}, masked pixel/patch prediction~\cite{pathak2016context,trinh2019selfie}, auto-regressive modeling~\cite{chen2020generative} , rotation prediction~\cite{gidaris2018unsupervised}, image colorization~\cite{zhang2016colorful,larsson2016learning}, cross-channel prediction~\cite{zhang2017split} and generative modeling~\cite{pu2016variational,donahue2019large}. These approaches typically exploit domain knowledge to carefully design a pretext task, with the learned features often focusing on one certain aspect of images, leading to a limited transfer ability.
$(\RN{2})$ 
{\em Contrastive learning.} The instance-level classification task is considered~\cite{dosovitskiy2015discriminative,zhuang2019local}, where each image in a dataset is treated as a unique class, and various augmented views of an image are the examples to be classified. Some recent works in this line are CPC~\cite{oord2018representation}, deep InfoMax~\cite{hjelm2018learning,bachman2019learning}, MoCo~\cite{he2020momentum}, SimCLR~\cite{chen2020simple}, BYOL~\cite{grill2020bootstrap} \etc.
$(\RN{3})$
{\em Prototype learning.} Clustering is employed for deep representation learning, including DeepCluster~\cite{caron2018deepcluster}, SwAV~\cite{caron2020unsupervised} and PCL~\cite{li2020prototypical}, among many others~\cite{xie2016unsupervised,yang2016joint,ji2019invariant,zhan2020online}. 
The proposed \shortname{} can be generally applied to all three classes in principle, as long as the notation of pseudo-labels exists. In this paper, we focus on the latter two classes, as they have shown SoTA representation learning performance, surpassing the ImageNet-supervised counterpart in multiple downstream vision tasks.

\paragraph{The role of augmentations.} Image data augmentations/transformations such as crop and blurring 
play a crucial role in modern self-supervised learning pipeline. It has been empirically shown that visual representations can be improved by employing stronger image transformations~\cite{chen2020improved} and increasing the number of augmented views of an image~\cite{caron2020unsupervised}. InfoMin~\cite{tian2020makes} studied the principles of good views for contrastive learning, and suggested to select views with less mutual information. By definition, adversarial and cut-mixed examples tend to be harder examples than transformation-augmented ones for self-supervised problems, and are complementary to the above techniques.

\subsection{Hard Examples}
\paragraph{Robustness.} A vast majority of works commonly view adversarial examples as a threat to models~\cite{goodfellow2014explaining,madry2017towards}, and suggest training with adversarial examples leads to accuracy drop on clean data~\cite{raghunathan2019adversarial,min2020curious}.
Adversarial training have been studied for self-supervised pre-training~\cite{chen2020adversarial}. Our work is significantly different from Chen \ea\cite{chen2020adversarial} in two aspects: 
$(\RN{1})$ Motivations -- We aim to use adversarial examples to boost standard recognition accuracy on large-scale datasets such as ImageNet, while Chen \ea\cite{chen2020adversarial} mainly study model robustness on small datasets such as CIFAR-10.
$(\RN{2})$ Algorithms -- We focus on the modern contrastive/prototype learning methods (last two categories of SSP methods in Section 4.1), while Chen \ea\cite{chen2020adversarial} work on traditional handcrafted SSP methods (the first category). 

\paragraph{Improved standard accuracy.} 
Hard examples have been shown to be effective in improving recognition accuracy in supervised learning settings.
For adversarial examples, one early attempt is virtual adversarial training (VAT)~\cite{miyato2018virtual}, a regularization method that improves semi-supervised learning tasks. The success was recently extended to natural language processing~\cite{wang2019improving,cheng2019robust,liu2020adversarial} and  vision-and-language tasks~\cite{gan2020large}. In computer vision, AdvProp~\cite{xie2020adversarial} is a recent work showing that  adversarial examples improve recognition accuracy on ImageNet in supervised settings. Hadi \ea further show that adversarially robust ImageNet models transfer better~\cite{salman2020adversarially}. For cut-mixed examples, it was first studied by Yun~\ea\cite{yun2019cutmix}. Similar augmentation schemes using a mixture of images include mix-up~\cite{zhang2017mixup}, cut-out~\cite{devries2017improved} \etc.
All above hard examples are constructed in the supervised settings, our \shortname{} is the first work to systematically study hard examples in large-scale self-supervised settings, due to the proposed pseudo-label formulation.  We confirm that hard examples improve the model's transfer ability.

\section{Experimental Results}

All of our study for unsupervised
pretraining (learning encoder network $f$ without labels)
is done using the ImageNet ILSVRC-2012 dataset~\cite{deng2009imagenet}. 
We implement \shortname{}$_{\text{MoCo}}$ based on the pre-training scheldule of MoCo-v2, and  implement \shortname{}$_{\text{Dcluster}}$ based on the pre-training scheldule of DeepCluster-v2. Both use the cosine learning rate and MLP projection head. Due to 
the limit of computational resource, all experiments are conducted with ResNet-50 and pre-trained in 200/800 epochs if not specifically mentioned. 
Once the model is pre-trained, we follow the same fine-tuning protocols/schedules with the baseline methods~\cite{he2020momentum,caron2020unsupervised}.
Following common practice in evaluating pre-trained visual representations, we test the model's transfer learning ability on a wide range of datasets/tasks in the self-supervised learning benchmark~\cite{goyal2019scaling}, based on the principle that a good representation should transfer with limited supervision
and limited fine-tuning.

\subsection{On the impact of different hard examples}
To understand different design choices in our framework, we compare different schemes to add hard examples into SSP.
$(\RN{1})$ {\em Std baseline}: Only standard random transformations are used, \ie, $\alpha_1=\alpha_2=0$;
$(\RN{2})$ {\em Std + Adv}: Adversarial examples are added into Std, \ie, $\alpha_1\!=\!1$ and $\alpha_2\!=\!0$;
$(\RN{3})$ {\em Std + Cmt}: Cutmixed examples are computed on $\Tilde{\xv}^{\text{Std}}$ and then added, \ie, $\alpha_1\!=\!0$ and $\alpha_2\!=\!1$;
$(\RN{4})$ {\em Std + Adv + Cmt}: Both hard examples are added, \ie,  $\alpha_1=\alpha_2=1$;
$(\RN{5})$ {\em Std + Adv + Cmt$_A$}: As an ablation choice, we consider computing cutmixed examples on adversarial views $\Tilde{\xv}^{\text{Adv}}$, denoted as {\em Cmx}$_A$;
$(\RN{6})$ {\em Std + Adv + Cmt + Cmt$_A$}: All types of hard examples are added. 

\begin{figure}[t!]
	\vspace{-0mm}\centering
	\begin{tabular}{c c}
		\hspace{-3mm}
		\includegraphics[height=5.5cm]{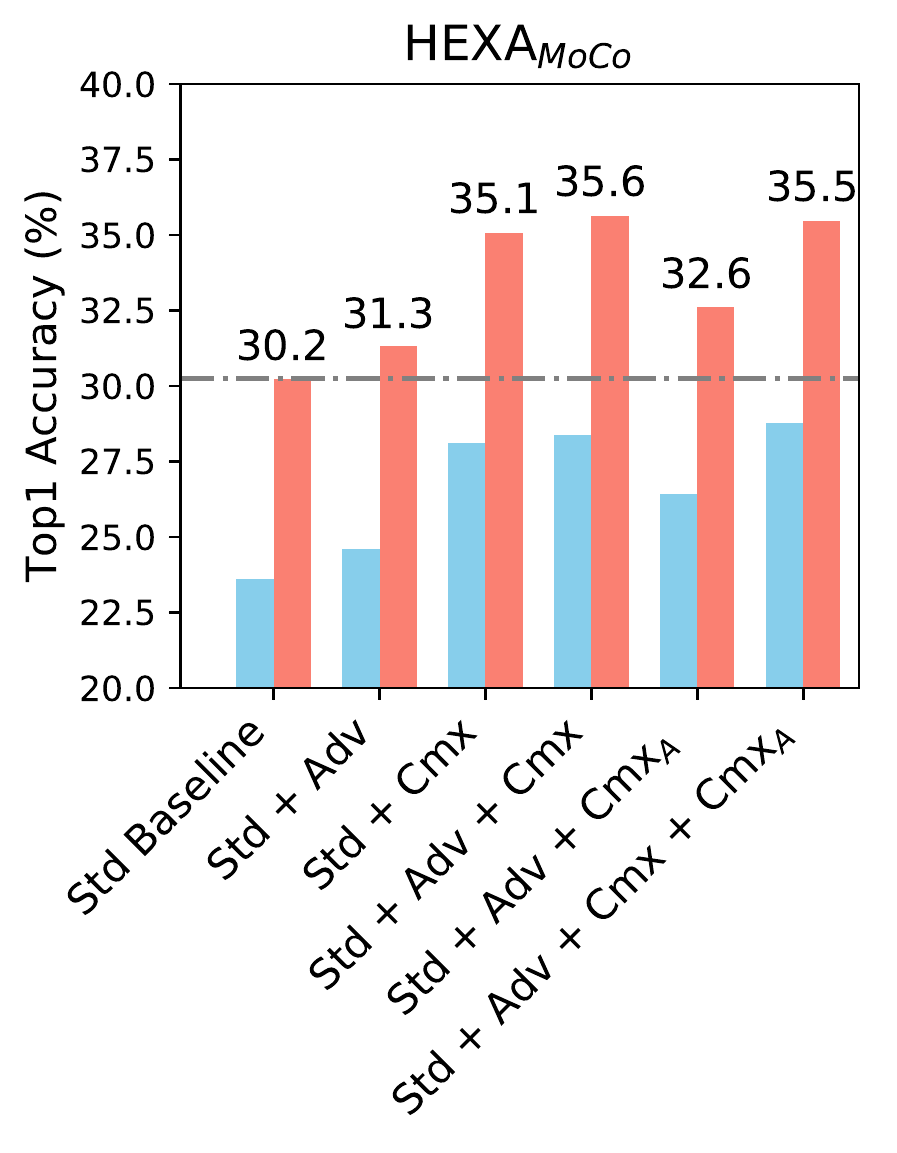}  & 
		\hspace{-6mm}
		\includegraphics[height=5.5cm]{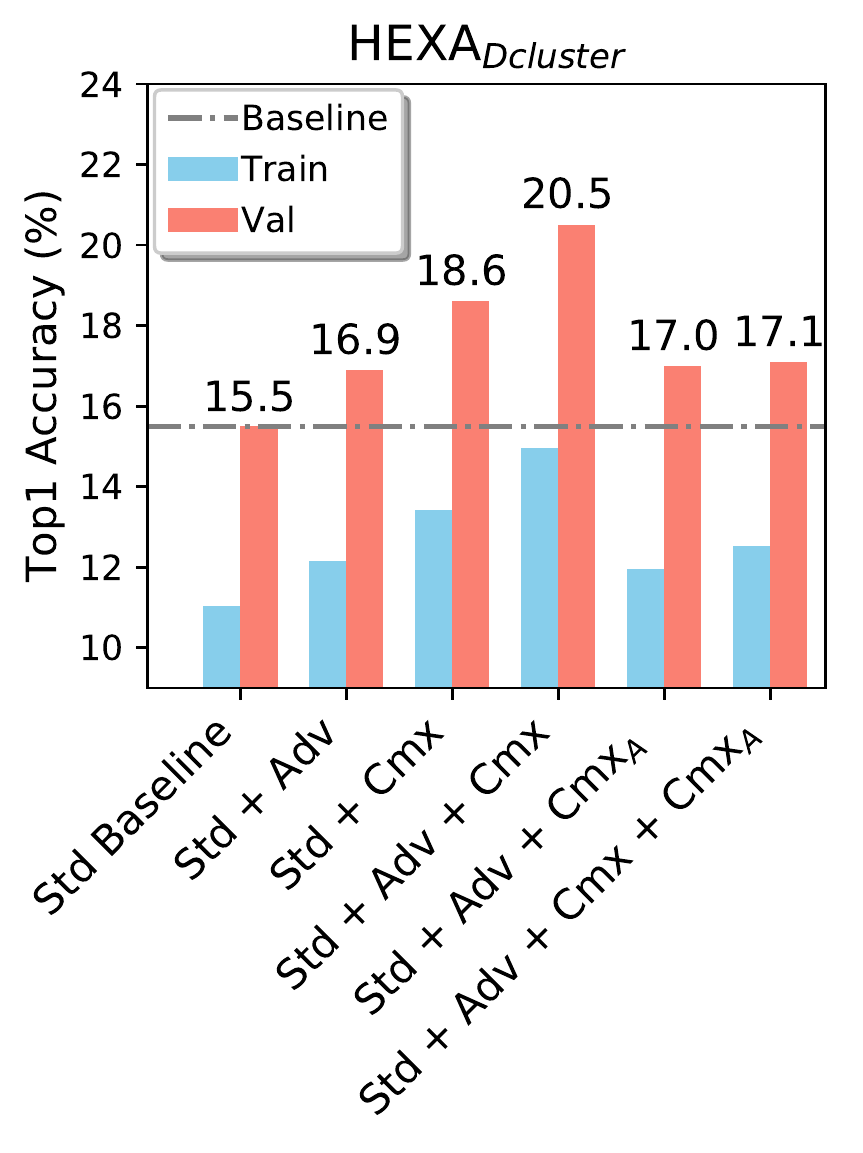} \\
		(a) \shortname{}$_{\text{MoCo}}$ \vspace{2mm} & 
		(b) \shortname{}$_{\text{Dcluster}}$  \hspace{-0mm} \\ 
	\end{tabular}
	\vspace{-4mm}
	\caption{Impact of different hard example combination schemes in \shortname{}.
	 }
	\vspace{-4mm}
	\label{fig:hard_example_impact}
\end{figure}

We conduct the comparison experiments with a small number of pre-training steps on ImageNet. For \shortname{}$_{\text{MoCo}}$, we pre-train for 20 epochs. For \shortname{}$_{\text{Dcluster}}$, we pre-train for 5 epochs, but with 6 crops per image: 2 crops at resolution 160 and 4 crops at resolution 96. The last checkpoint is employed to extract features, on which a linear classifier is trained for 1 epoch on ImageNet. The results are reported in Figure~\ref{fig:hard_example_impact}. 
Interestingly, cut-mixed examples computed on $\Tilde{\xv}^{\text{Std}}$  are more effective than those on $\Tilde{\xv}^{\text{Adv}}$. This is expected, as the ground-truth label of $\Tilde{\xv}^{\text{Adv}}$ should be different from $\Tilde{\xv}^{\text{Std}}$, the mixed label of the latter can not reflect ground-truth label of the former. Further, both adversarial and cut-mixed examples can improve the baseline method, regardless of whether they are added separately or simultaneously, showing the effectiveness of the proposed methods.

In what follows, 
we denote \shortname{}$_{\text{MoCo}}$ and \shortname{}$_{\text{Dcluster}}$ as two variants that are both 
constructed with 2 random crops at resolution 224 and adversarial examples. 
More specifically, \shortname{}$_{\text{MoCo}}$ follows MoCo-v2: one crop for query and the other for key; \shortname{}$_{\text{Dcluster}}$ is always compared with the DeepCluster-v2 variant with 2 crops. 
The current SoTA method is SwAV~\cite{caron2020unsupervised}, which employs 8 random crops: 2 crops at resolution 224 and 6 crops at resolution 96. To compare with SoTA,  we also increase the number of crops to 8 and consider two variants: \shortname{}$_{\text{Dcluster}}$(8-crop) is with adversarial examples, and \shortname{}$_{\text{Dcluster}}^{+}$(8-crop) is constructed with both adversarial and cut-mixed examples. 
All 2-crop methods use a mini-batch size of B=256, and 8-crop methods use a mini-batch size of B=4096.

\subsection{Linear classification}
To evaluate the learned representations, we first follow the widely used
linear evaluation protocol, where a linear classifier is trained on top of the frozen base network, and test accuracy is used as a proxy for representations. We follow previous setup~\cite{goyal2019scaling} and evaluate the performance of such linear classifiers on four datasets, including ImageNet~\cite{deng2009imagenet}, PASCAL VOC2007 (VOC07)~\cite{everingham2010pascal}, CIFAR10 (C10) and CIFAR100 (C100)~\cite{krizhevsky2009learning}.  A softmax classifier is trained for ImageNet/CIFAR, while a linear SVM~\cite{fan2008liblinear} is trained for VOC07. We report 1-crop ($224\times224$), Top-1 validation accuracy for ImageNet/CIFAR and mAP for VOC07.

\begin{table}[t!]
\footnotesize 
\centering
\begin{tabular}{ @{\hspace{-0pt}}l@{\hspace{8pt}}c@{\hspace{2pt}}|@{\hspace{7pt}}c@{\hspace{7pt}}c@{\hspace{7pt}}c@{\hspace{7pt}}c}
\toprule
 Method   & Epoch &  ~~ImageNet  &  VOC07 &  C10 &  C100  \\ 
\hline
Supervised & - & 76.5 &  87.5 & 93.6 & 78.3 \\
\hline
Instance D.~\cite{wu2018unsupervised} & 200 & 54.0 &  - & -  & -\\ 
Jigsaw~\cite{noroozi2016unsupervised} & 90 & 45.7 & 64.5 & - & -\\ 
BigBiGAN~\cite{donahue2019large} & - & 56.6 & - &  - & -\\
CPC-v2~\cite{henaff2019data} & 200 & 63.8 & - & - & -\\
CMC~\cite{tian2019contrastive} & 200 &  66.2 & - & - & -\\
SimCLR~\cite{chen2020simple} & 200 & 61.9 & -  & - & -\\
SimCLR~\cite{chen2020simple} & 1000 & 69.3 & 80.5 & 90.6 & 71.6 \\
MoCo~\cite{he2020momentum} & 200 & 60.6 & 79.2 & - & - \\ 
PIRL~\cite{misra2020self} & 800 & 63.6 & 81.1 & - & - \\ 
PCL-v2~\cite{li2020prototypical} & 200 & 67.6 & 85.4 & - & \\
BYOL~\cite{grill2020bootstrap} & 800 & 74.3 & - & 91.3 & 78.4 \\
SwAV \!(B=256)\!~\cite{caron2020unsupervised} & 200 & 72.7 & 87.5 & 91.8 & 74.2 \\
SwAV \!(B=4096)\!~\cite{caron2020unsupervised} & 200 & 73.9 &  87.9 & 92.0 & 76.0 \\
SwAV \!(B=4096)\!~\cite{caron2020unsupervised} & 800 & 75.3 &  88.1 & 93.1 & 77.0 \\
InfoMin~\cite{tian2020makes} & 200 & 70.1 & - & - & -\\
InfoMin~\cite{tian2020makes} & 800 & 73.0 & - & - & - \\
\hline
 MoCo-v2~~ &  200  & 67.5 & 84.5 & 89.4 & 70.1\\
\rowcolor{Gray}
\cellcolor{white}
   \shortname{}$_{\text{MoCo}}$  &  200   & \textcolor{blue}{68.9} & \textcolor{blue}{85.0} & \textcolor{blue}{90.4} &  \textcolor{blue}{71.5}
  \\ \hline
 MoCo-v2~~ &  800  & 71.1 &  86.8  & 90.6 & 71.8 \\
\rowcolor{Gray}
\cellcolor{white}
   \shortname{}$_{\text{MoCo}}$  &  800   & \textcolor{blue}{71.7} &  \textcolor{blue}{87.0} & \textcolor{blue}{91.5}  & \textcolor{blue}{73.2} 
  \\ \hline
 DeepCluster-v2~~ &  200  & 67.6 &  85.4 & 89.6 & 70.9\\
\rowcolor{Gray}
\cellcolor{white}
   \shortname{}$_{\text{DCluster}}$  &  200   & \textcolor{blue}{68.1} & 
 \textcolor{blue}{85.9}  &    \textcolor{blue}{90.7} &  \textcolor{blue}{71.5}  \\  
\rowcolor{Gray}
\cellcolor{white}
   \shortname{}$_{\text{DCluster}}$(8-crop \!)  &  200   & \textcolor{blue}{\bf{74.0}} & 
  \textcolor{blue}{88.1} &   \textcolor{blue}{ \bf{92.9}} &  \textcolor{blue}{\bf{76.6}}    \\  
\rowcolor{Gray}
\cellcolor{white}
   \shortname{}$_{\text{DCluster}}^{+}$(8-crop \!)   &  200   & \textcolor{blue}{73.4} & 
   \textcolor{blue}{ \bf{88.8}}  &   \textcolor{blue}{91.9}  &  \textcolor{blue}{75.2}     \\ \hline
 DeepCluster-v2 (8-crop \!) &  800  & 75.2 &  87.6 & 93.2 & 77.3\\ 
 \rowcolor{Gray}
\cellcolor{white}
   \shortname{}$_{\text{DCluster}}$(8-crop \!)   &  800   & \textcolor{blue}{\bf{75.5}} & 
   \textcolor{blue}{87.9}  &   \textcolor{blue}{93.4}  &  \textcolor{blue}{\bf{78.6}}     \\ 
\rowcolor{Gray}
\cellcolor{white}
   \shortname{}$_{\text{DCluster}}^{+}$(8-crop \!)   &  800   & \textcolor{blue}{75.1} & 
   \textcolor{blue}{\bf{88.2}}  &   \textcolor{blue}{\bf{93.5}}  &  \textcolor{blue}{78.0}     \\
\bottomrule
\end{tabular}

\vspace{-1mm}
\caption{Linear classification performance on learned representations using ResNet-50. All numbers for baselines are from their corresponding papers or~\cite{li2020prototypical} , except that we use the released pretrained model for SwAV.}
\label{tab:main_result_cls}
\vspace{-0mm}
\end{table}

\begin{figure}[t!]
	\vspace{-3mm}\centering
	\begin{tabular}{cc}
	    \hspace{-4mm}		
		\includegraphics[height=3.6cm]{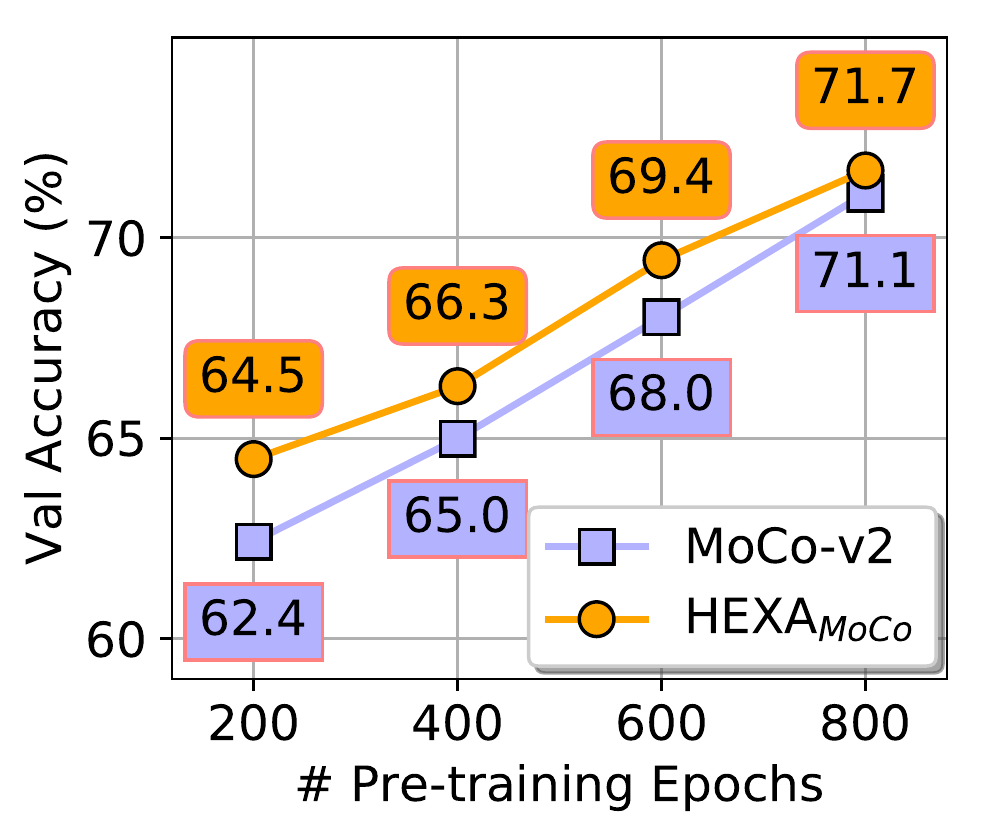} & \hspace{-6mm}
		\includegraphics[height=3.6cm]{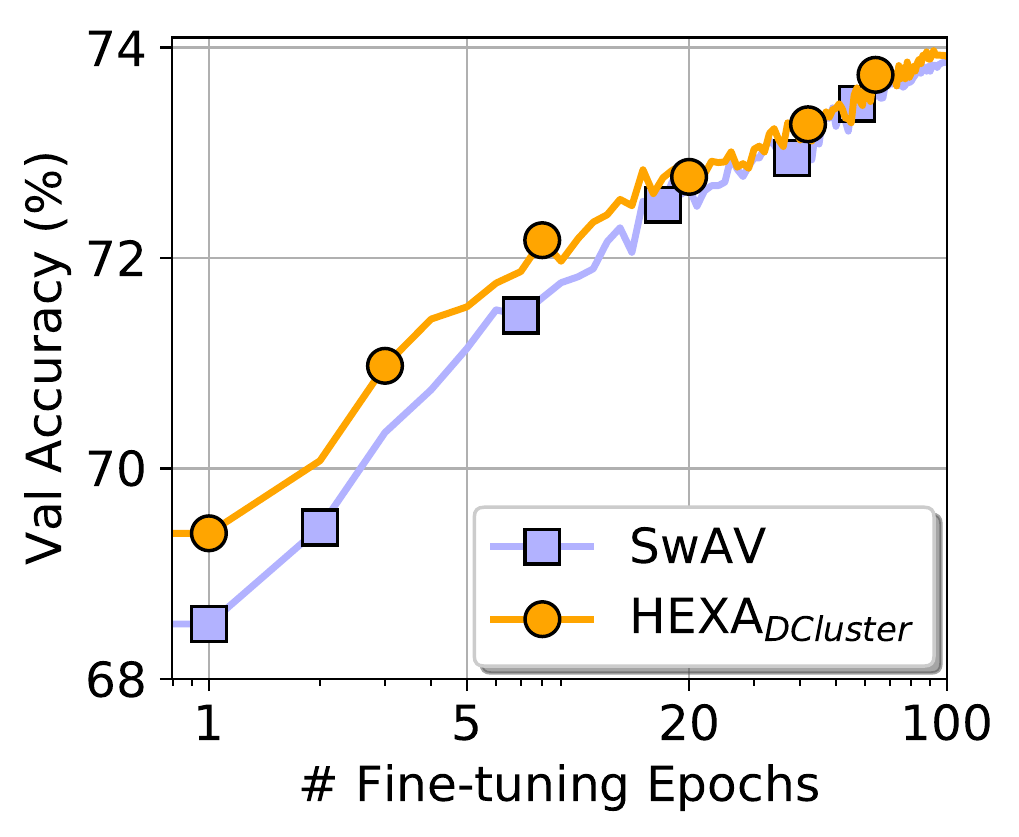} \\
		(a) Pre-training & (b) Linear classification
	\end{tabular}
	\vspace{-2mm}
	\caption{Learning curves on ImageNet. (a) For 800 pre-training epochs of contrastive methods, the Top-1 accuracy is measured for checkpoints at every 200 epochs. (b) Training a linear classifier on the 200th checkpoint produced by prototype methods for 100 epochs.}
	\vspace{-4mm}
	\label{fig:learning_curves}
\end{figure}

Table~\ref{tab:main_result_cls} shows the results of linear classification. It is interesting to observe that DeepCluster-v2 is slightly better than MoCo-v2, indicating that the traditional prototype methods can be on par with the popular contrastive methods, with the same pre-training epochs and data augmentation strategies. We hope this result can inspire future research to more carefully select different pretext objectives. 
By contrast, \shortname{} variants consistently outperform theirs counterparts for both contrastive and prototype methods, demonstrating that the proposed hard examples can effectively improve learned visual representations in SSP. 

We also pre-train \shortname{}$_{\text{MoCo}}$ with 800 epochs, a longer schedule used in MoCo-v2~\cite{chen2020improved}. The learning curves are compared in Figure~\ref{fig:learning_curves}(a). 
\shortname{} is consistently better than MoCo-v2 and the gap is larger at the beginning. We hypothesize that the augmentation space is more efficiently explored with hard examples than with traditional image transformations, but this advantage is less reflected in improved recognition accuracy, when the augmentation space is gradually fully occupied at the end of training.
When comparing with SoTA methods equipped with multi-crop~\cite{caron2020unsupervised}, we see that \shortname{}$_{\text{Dcluster}}$(8-crop) achieves slightly better than SwAV on ImageNet, and even outperforms InfoMin with 800 pre-training steps. By plotting the training curves of their linear classifiers in Figure~\ref{fig:learning_curves}(b), we observe that \shortname{}$_{\text{Dcluster}}$(8-crop) clearly outperforms SwAV with limited fine-tuning (\eg $<$20 epochs training). The advantage of \shortname{} is more significantly than SwAV with limit supervision, this can be seen from a larger performance gap on VOC07 in Table~\ref{tab:main_result_cls}.







\begin{table}[t!]
\footnotesize 
\centering
\begin{tabular}{@{\hspace{-0pt}}l@{\hspace{8pt}}c@{\hspace{5pt}}|c@{\hspace{7pt}}c@{\hspace{7pt}}c@{\hspace{7pt}}c@{\hspace{7pt}}c }
\toprule
 Method   & Epoch &  2  & 4 &  8 & 16 & 32\\ 
\hline
 Supervised~~ &  -  &  67.8 & 73.9 & 79.6 & 82.3 & 83.8  \\
\hline
Jigsaw~\cite{noroozi2016unsupervised} & 200 & 31.1 & 40.0 & 46.7 & 51.8 & - \\
SimCLR~\cite{chen2020simple} & 200 &   43.1 & 52.5 & 61.0 & 67.1 & - \\
MoCo~\cite{he2020momentum} & 200 &  42.0 & 49.5 & 60.0 & 65.9 &  -  \\        
PCL-v2~\cite{li2020prototypical} & 200 & 59.6  & 66.2 &  74.5  & 78.3 &  -\\
SwAV \!(B=4096)\!~\cite{caron2020unsupervised} & 200 & 53.8 & 65.0 & 73.9 &  78.6 &  82.3  \\
SwAV \!(B=4096)\!~\cite{caron2020unsupervised} & 800 & 54.6 & 64.6 & 73.4 &  79.0 &  82.5  \\
\hline
 MoCo-v2~\cite{chen2020improved} &  200  & 56.4  & 67.2  & 72.0  & 77.2 &  79.5 \\
\rowcolor{Gray}
\cellcolor{white}
   \shortname{}$_{\text{MoCo}}$  &  200   & \textcolor{blue}{57.0}   & \textcolor{blue}{68.5}   &  \textcolor{blue}{73.1}   & \textcolor{blue}{78.0}  & \textcolor{blue}{80.1}
  \\ \hline 
 MoCo-v2~\cite{chen2020improved} &  800  & 60.6 & 72.1 & 77.1 & 80.9 & 82.8 \\
\rowcolor{Gray}
\cellcolor{white}
   \shortname{}$_{\text{MoCo}}$  &  800   & \textcolor{blue}{\bf{61.5}}   & \textcolor{blue}{\bf{72.8}}   &  \textcolor{blue}{\bf{77.5}}   & \textcolor{blue}{\bf{81.5}}  & \textcolor{blue}{83.0}
  \\ \hline 
  DeepCluster-v2~\cite{caron2018deepcluster,caron2020unsupervised} &  200  &  
  57.7 &  66.5 & 74.1 & 77.6 & 80.7 \\  %
 \rowcolor{Gray}
\cellcolor{white}
  \shortname{}$_{\text{DCluster}}$  & 200  & 55.7 & 65.3 & 74.1 & \textcolor{blue}{78.0} & \textcolor{blue}{81.2}\\
 \rowcolor{Gray}
\cellcolor{white}
  \shortname{}$_{\text{DCluster}}$(8-crop \!)   &  200 &  55.5 & 66.2 & \textcolor{blue}{75.2} & \textcolor{blue}{79.5} & \textcolor{blue}{83.1} \\
\rowcolor{Gray}
\cellcolor{white}
   \shortname{}$_{\text{DCluster}}^{+}$(8-crop \!)  &  200   & 56.9  & \textcolor{blue}{67.3}   & \textcolor{blue}{\bf{76.4}} & \textcolor{blue}{\bf{81.1}}  & \textcolor{blue}{\bf{84.0}}
  \\  \hline 
  DeepCluster-v2~\cite{caron2018deepcluster,caron2020unsupervised} &  800  &  
  53.5 &  65.6 & 73.3 & 78.9 & 82.5 \\  
 \rowcolor{Gray}
\cellcolor{white}
  \shortname{}$_{\text{DCluster}}$(8-crop \!)   &  800 &  \textcolor{blue}{54.1} & 65.6 & \textcolor{blue}{73.9} & \textcolor{blue}{79.0} & \textcolor{blue}{82.9} \\  
 \rowcolor{Gray}
\cellcolor{white}
  \shortname{}$_{\text{DCluster}}^{+}$(8-crop \!)   &  800 &  \textcolor{blue}{54.8} & \textcolor{blue}{65.8} & \textcolor{blue}{74.2} & \textcolor{blue}{79.4} & \textcolor{blue}{83.1} \\ 
\bottomrule
\end{tabular}

\vspace{-1mm}
\caption{Low-shot classification on VOC07 using linear SVMs trained on
fixed representations. We vary the number of labeled examples $k$ per class and report the mAP across 5 runs. All baseline numbers are from~\cite{li2020prototypical} except that we use the released pretrained model for SwAV.}
\label{tab:low_shot}
\vspace{-3mm}
\end{table}

\paragraph{Low-shot classification.} We evaluate the learned representation on image classification tasks with few training samples per-category. We follow the setup in Goyal \ea\cite{goyal2019scaling} and train linear SVMs using fixed representations on VOC07
for object classification. We vary the number $k$ of training samples per-class and report the average result
across 5 independent runs. The results are shown in Table~\ref{tab:low_shot}. Hard examples help improve the performance for both contrastive and prototype learning, especially when $k \ge 8$. This is probably because the performance is very sensitive to the choice of selected labelled samples when $k \le 4$, rendering the evaluation less stable.  Pre-training longer (MoCo-v2 with 800 epochs) helps reduce this issue, and the proposed hard examples can further boost the performance. When $k\ge32$ samples are considered, the proposed scheme surpasses the ImageNet-supervised pre-training approach. To the best of our knowledge, \shortname{} is the first work to surpasses the supervised baseline with such a small number of labelled samples on VOC07, showing high sample-efficiency of the learned representations. \shortname{} pre-trained at 200 epochs also outperforms SwAV (pre-trained at both 200 epochs and 800 epochs) by a large margin in all cases.

\begin{table}[t!]
\scriptsize
\centering
\begin{tabular}{@{}lp{18pt}|p{18pt}@{}p{18pt}@{}|p{18pt}@{}p{18pt}|p{18pt}@{}p{18pt}}
\toprule
&   & \multicolumn{2}{c|}{1\% labels} & \multicolumn{2}{c|}{10\% labels} & \multicolumn{2}{c}{100\% labels} \\
 Method   & Epoch & Top-1 &  Top-5  & Top-1 &  Top-5 & Top-1 &  Top-5  \\ 
\hline
 Supervised~~ &  -  &  25.4 & 48.4 & 56.4 & 80.4 & 76.5 & 93.0\\
\hline
\multicolumn{6}{l}{{\hspace{-3mm} \em Semi-supervised: }}  \\
\hline
Pseudolabels~\cite{zhai2019s4l} &  - & - &  51.6 & - &82.4 & - & -\\
VAT ~\cite{miyato2018virtual,zhai2019s4l} & -  & - & 47.0 & - & 83.4 & - & -\\
S$^4$L Rotation~\cite{zhai2019s4l} & - & - &  53.4 & - & 83.8 & - & -\\
UDA~\cite{xie2019unsupervised}  & - & - & - & 68.8 & 88.5 & - & -\\
FixMatch & - & - & - & 71.5  & 89.1 & - & -\\
   \hline
\multicolumn{6}{l}{{\hspace{-3mm} \em Self-supervised: }}  \\
\hline
Instance D.~\cite{wu2018unsupervised} & 200 &  - &   39.2 & - & 77.4& - & - \\
Jigsaw~\cite{noroozi2016unsupervised} & 90 & - & 45.3 & - &  79.3 & - & -\\
SimCLR~\cite{chen2020simple} & 200 & - & 56.5 & - & 82.7 & - & -\\
SimCLR~\cite{chen2020simple} & 1000 & 48.3 & 75.5 & 65.6  & 87.8 & 76.5 & 93.5\\
MoCo~\cite{he2020momentum} & 200 & - & 56.9 & - & 83.0 & - & -\\        
PIRL~\cite{misra2020self} & 800 & - & 57.2 & - & 83.8 & - & -\\
PCL~\cite{li2020prototypical} & 200 & - & 75.3 & - & 85.6 & - & -\\
SWAV \!(B=256)\!~\cite{caron2020unsupervised} & 200 & 51.3 & 76.6 & 67.8 & 88.6  & 75.5 & 92.9 \\
SWAV \!(B=4096)\!~\cite{caron2020unsupervised} & 200 & 52.6 & 77.7 & 68.5 & 89.2  & 76.3 & 93.2 \\
SWAV \!(B=4096)\!~\cite{caron2020unsupervised} & 800 & 53.9 & 78.5 & 70.2 & 89.9 & 78.3 & 94.1\\ 
BYOL~\cite{grill2020bootstrap} & 800 & 53.2 & 78.4 & 68.8 & 89.0 & 77.7 & 93.9\\
\hline
 MoCo-v2~\cite{chen2020improved} &  200  & 38.9  & 67.4  &  61.5 & 84.6
 & 74.6  & 92.5 \\
\rowcolor{Gray}
\cellcolor{white}
   \shortname{}$_{\text{MoCo}}$  &  200   & \textcolor{blue}{39.4}   & \textcolor{blue}{67.6}   &  \textcolor{blue}{62.3}   & \textcolor{blue}{85.1}  & 
   \textcolor{blue}{74.8}   & \textcolor{blue}{92.4} 
  \\ \hline 
 MoCo-v2~\cite{chen2020improved} &  800  & 42.3 &  70.1  & 63.8 &  86.2 & 75.5 & 92.8\\
\rowcolor{Gray}
\cellcolor{white}
   \shortname{}$_{\text{MoCo}}$  &  800   & \textcolor{blue}{42.4}  & \textcolor{blue}{70.2}  &  \textcolor{blue}{64.1} & \textcolor{blue}{86.3} & 
    \textcolor{blue}{75.7} & \textcolor{blue}{93.0}
  \\ \hline
 DeepCluster-v2~\cite{caron2018deepcluster,caron2020unsupervised} &  200  & 46.7 & 72.9  & 63.5 & 86.3 & 71.9 &  91.0\\
\rowcolor{Gray}
\cellcolor{white}
   \shortname{}$_{\text{DCluster}}$  &  200   & \textcolor{blue}{48.9}  & \textcolor{blue}{74.7}   & \textcolor{blue}{64.9} & \textcolor{blue}{87.3} & 
   \textcolor{blue}{ 73.9} & \textcolor{blue}{92.2} 
  \\  
\rowcolor{Gray}
\cellcolor{white}
   \shortname{}$_{\text{DCluster}}$ (8-crop)  &  200   & \textcolor{blue}{54.1}  &  \textcolor{blue}{78.6} & \textcolor{blue}{69.3}  &  \textcolor{blue}{89.3} & 
\textcolor{blue}{76.9}  &  \textcolor{blue}{93.6}    
  \\    
\rowcolor{Gray}
\cellcolor{white}
  \shortname{}$_{\text{DCluster}}^{+}$ (8-crop)  &  200 &   \textcolor{blue}{54.9} & \textcolor{blue}{79.3} & \textcolor{blue}{69.4} & \textcolor{blue}{89.7} &
  \textcolor{blue}{77.2}  &  \textcolor{blue}{93.9} 
  \\ \hline
 DeepCluster-v2~\cite{caron2018deepcluster,caron2020unsupervised} &  800  & 55.6 & 79.3  & 70.6 & 90.2 & 78.0 & 94.0 
  \\  
\rowcolor{Gray}
\cellcolor{white}
   \shortname{}$_{\text{DCluster}}$ (8-crop)  &  800   & 55.3  & 79.1 & \textcolor{blue}{70.8}  &  \textcolor{blue}{90.2} & 
\textcolor{blue}{78.3}  &  \textcolor{blue}{94.1}    
\\
\rowcolor{Gray}
\cellcolor{white}
  \shortname{}$_{\text{DCluster}}^{+}$ (8-crop)  &  800 &   \textcolor{blue}{\textbf{57.3}} & \textcolor{blue}{\textbf{80.7}} & \textcolor{blue}{\textbf{71.8}} & \textcolor{blue}{\textbf{90.8}} & 
  \textcolor{blue}{\textbf{78.6}} & \textcolor{blue}{\textbf{94.4}} \\ 
  

\bottomrule
\end{tabular}

\vspace{-1mm}
\caption{Semi-supervised classification on ImageNet. We use the released pretrained model
for MoCo/SwAV. All other numbers are adopted from corresponding papers.}
\label{tab:semisup}
\vspace{-4mm}
\end{table}

\subsection{Semi-supervised learning on ImageNet}

We perform semi-supervised learning experiments to evaluate whether the learned representation can provide a good basis for fine-tuning. Following the setup from Chen \ea\cite{chen2020simple}, we select a subset (1\% or 10\%) of ImageNet training data (the same labelled images with Chen\ea\cite{chen2020simple}),
and fine-tune the entire self-supervised trained model on these subsets.
For the proposed \shortname{}, and  we fine-tune the models using the same schedule. SwAV with 8 augmentation crops and 200 pre-training epochs is used a fair baseline.

Table~\ref{tab:semisup} reports the Top-1 and Top-5 accuracy on ImageNet validation set. \shortname{} improves its counterparts MoCo-v2 and DeepCluster-v2 in all cases. By different variants of  \shortname{}$_{\text{DCluster}}$, we see that cut-mixed examples are important in boosting performance, especially with 1\% labels. 
\shortname{}$_{\text{DCluster}}^{+}$(8-crop \!) sets a new SoTA under 200 training epochs, outperforming all existing self-supervised learning methods. 
It even outperforms BYOL pre-trained at 800 epochs in both cases.
For SwAV pre-trained at 200 epochs, it is significantly inferior to \shortname{} in the same setting.
For SwAV pre-trained at 800 epochs, it achieves Top-1 53.9\% and Top-5 78.5\% with 1\% labelled images, which is lower than our \shortname{} pre-trained at 200 epochs by a notable margin. This again shows the effectiveness of hard examples in improving visual representations in low-resource settings. 

We also fine-tune over 100\% of ImageNet labels for 20 epochs, and \shortname{} reaches 78.6\% Top-1 accuracy, outperforming the supervised approach (76.5\%) using the same ResNet-50 architecture by a large margin (2.1\% absolute recognition accuracy). \shortname{} also achieves higher performance compared with all existing self-supervised learning methods in both 200 and 800 pre-training epochs settings. This shows that hard examples can effectively improve SSP, which can be viewed as a promising approach to further improve standard supervised learning such as Big Transfer~\cite{kolesnikov2019big} in the future.




\begin{wraptable}{R}{0.25\textwidth}
\vspace{-5mm}
\begin{minipage}{0.25\textwidth}
\scriptsize
\centering
\hspace{-3mm}
\begin{tabular}{@{\hspace{3pt}}l@{\hspace{5pt}}l@{\hspace{1pt}} |@{\hspace{6pt}}c@{\hspace{6pt}}c@{\hspace{5pt}}c}
\hline
Methods &  Epoch  & AP & AP50 & AP75 \\ 
\hline
Supervised & - & ~53.5 & 81.3 & 58.8 \\ \hline
MoCo-v2 & 200  & ~57.0 &  82.4 & 63.6 \\ 	
\rowcolor{Gray}
\cellcolor{white}
\shortname{}$_{\text{MoCo}}$ & 200 &  ~\textcolor{blue}{57.1}  & \textcolor{blue}{82.4}  &  \textcolor{blue}{63.8} 
  \\ \hline
MoCo-v2 &  800 ~~  & ~57.4 & 82.5 & 64.0  \\
\rowcolor{Gray}
\cellcolor{white}
 \shortname{}$_{\text{MoCo}}$ &  800 & ~\textcolor{blue}{\textbf{57.7}} &  \textcolor{blue}{\textbf{82.8}}  &  \textcolor{blue}{\textbf{64.9}} \\ 
\bottomrule
 \end{tabular}
\vspace{-1mm}
\caption{\small Object detection results on VOC. The numbers for MoCo-v2 are from~\cite{chen2020improved}.} 
\label{tab:obj_detection}
\vspace{-3mm}
\end{minipage}
\end{wraptable}





\subsection{Object detection}
It is standard practice in data-scarce object detection tasks to initialize earlier
model layers with the weights from ImageNet-trained networks. We study the benefits of using hard-examples-trained networks to initialize object detection. On the VOC object detection task, a Faster R-CNN
detector~\cite{ren2015faster} is fine-tuned end-to-end on the VOC 07+12 trainval set1 and evaluated on the VOC 07 test set using the COCO suite of metrics~\cite{lin2014microsoft}. The results are shown in Table~\ref{tab:obj_detection}. We find that \shortname{} consistently outperforms MoCo-v2 that is pre-trained with standard image transformations. 

\section{Conclusion}
We have presented a comprehensive study of utilizing hard examples to improve visual representations for image self-supervised learning. By treating SSP as a pseudo-label classification task, we introduce a general framework to generate {\em harder} augmented views to boost the discriminative power of self-supervised learned models. Two novel algorithmic variants are proposed: \shortname{}$_{\text{MoCo}}$ for contrastive learning and \shortname{}$_{\text{DCluster}}$ for prototype learning. Our \shortname{} variants outperform their counterparts, often by a notable margin, and achieve SoTA under the same settings. Future research directions include incorporating more advanced hard examples under this framework, and exploring their performance with larger networks.

\newpage

\paragraph{Acknowledgments}  The authors gratefully acknowledge Bai Li for
helpful discussion. Additional thanks go to the entire Project Philly team inside Microsoft, who provided us the computing platform for our research.

{\small
\bibliographystyle{ieee_fullname}
\bibliography{egbib}
}

\appendix
\newpage




%
\section{Hyper-parameter Choice}

\begin{figure}[h!]
	\vspace{-0mm}\centering
	\begin{tabular}{c}
	    \hspace{-4mm}		
		\includegraphics[height=5.2cm]{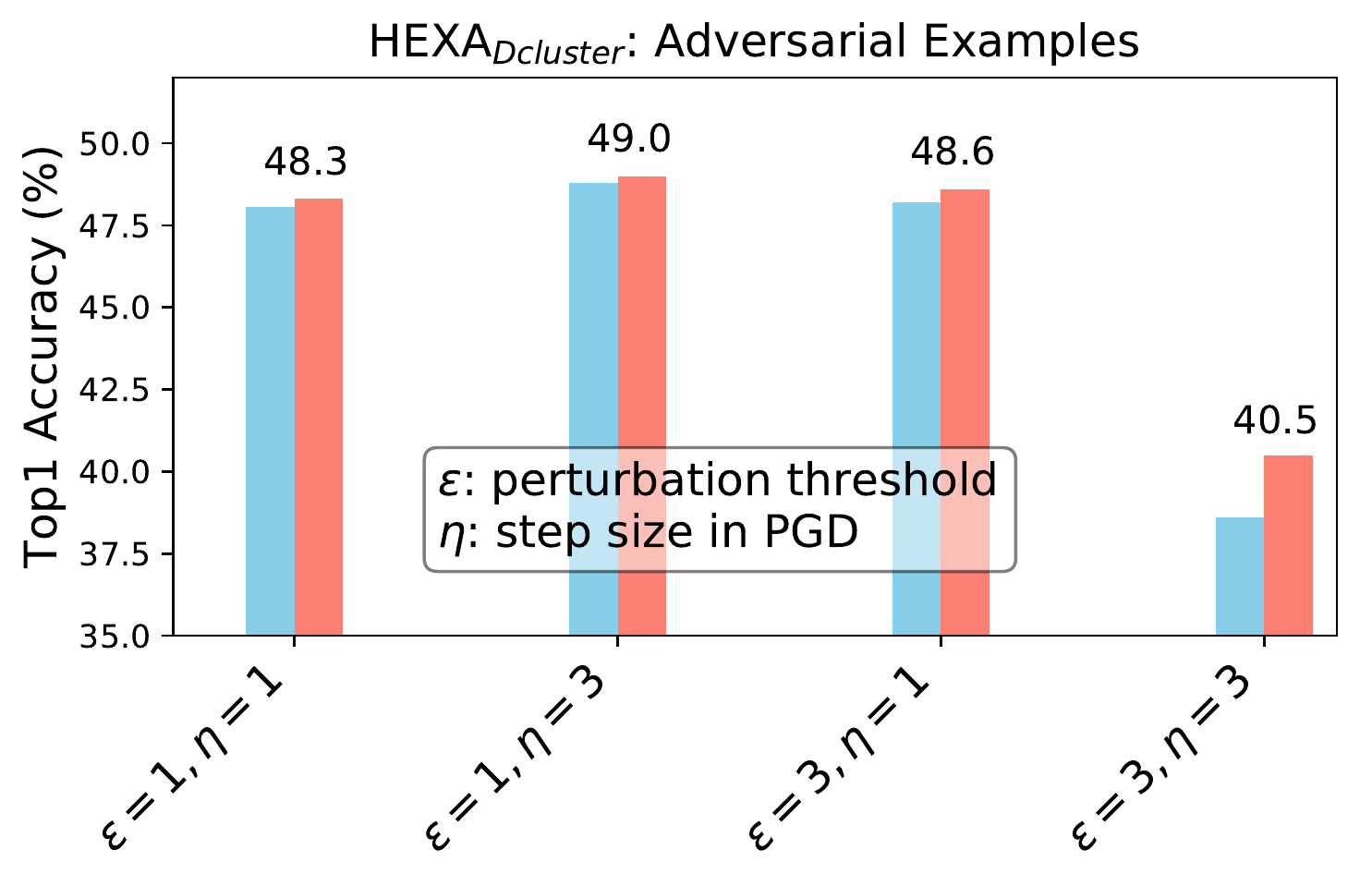}
	\end{tabular}
	\vspace{-2mm}
	\caption{Top-1 accuracy on ImageNet is measured for different adversarial attack settings.}
	\vspace{-2mm}
	\label{fig:adv_example_ablation}
\end{figure}

\paragraph{Adversarial examples.}
We study the hyper-parameter choices attack perturbation threshold $\epsilon$ and PGD step size $ \eta $  in adversarial images. For \shortname{}$_{\text{DCluster}}$, we grid search over $\epsilon = [1,3]$ and $ \eta  = [1,3]$. Each variant is pre-trained for 40 epochs. A linear classifier is added on the  pre-trained checkpoint and trained for 100 epochs. The results are shown Figure~\ref{fig:adv_example_ablation}. Adding too large perturbations  $(\epsilon = 3,  \eta  = 3)$ can hurt model performance significantly. Otherwise, the model perform similarly with differnt ways of adding small perturbations, allowing either a large threshold with a small step size, or a large step size with a small threshold. We used $(\epsilon = 1, \eta = 1)$ for convenience.

\begin{figure}[h!]
	\vspace{-0mm}\centering
	\begin{tabular}{c}
	    \hspace{-4mm}		
		\includegraphics[height=3.0cm]{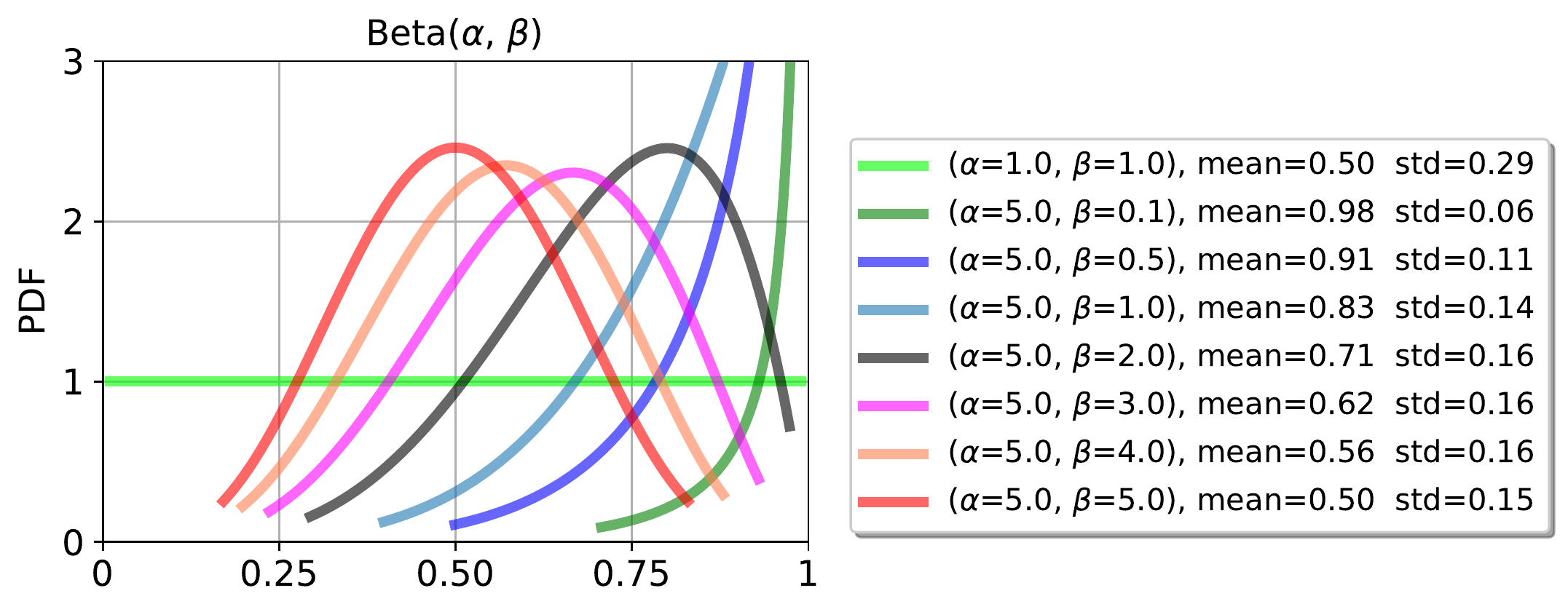} \\
		(a) PDF of Beta distribution \\
	    \includegraphics[height=5.2cm]{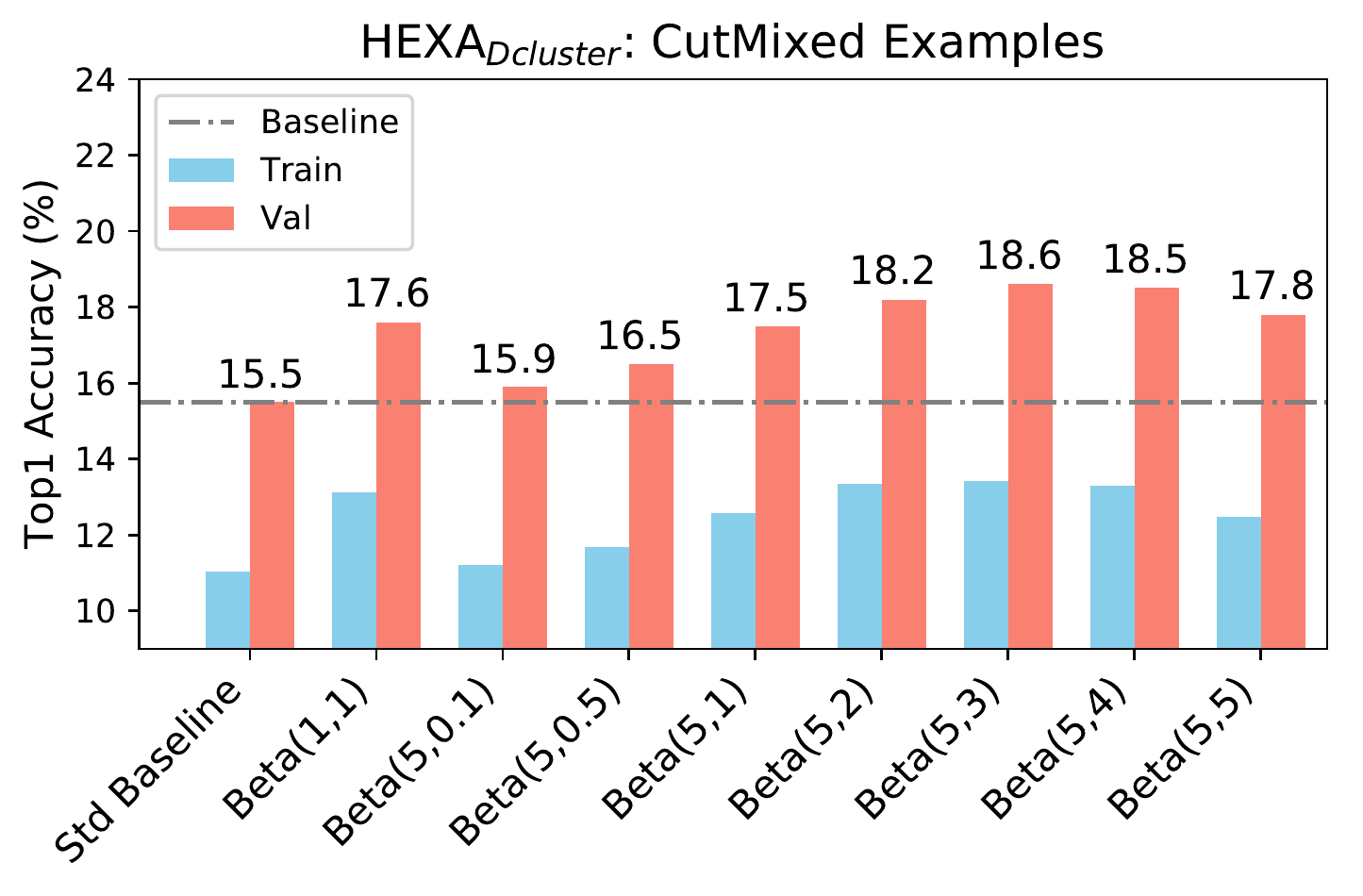} \\
		(b) Top-1 accuracy on ImageNet
	\end{tabular}
	\vspace{-2mm}
	\caption{The impact of hyper-parameters in mixing two images, measured by Top-1 accuracy on ImageNet. (a) The probability density function (PDF) of a Beta distribution. (b) Accuracy of \shortname{}$_{\text{DCluster}}$ checkpoints with 5 pre-training epochs.}
	\vspace{-2mm}
	\label{fig:cutmix_example_ablation}
\end{figure}

\paragraph{Cut-mixed examples.}
We study the hyper-parameter choices in $\Beta(\alpha, \beta)$ in cut-mixed images. We consider 6 random crops for each image: 2 crops at resolution 160 and 4 crops at resolution 96. The model is pre-trained in 5 epoch, and a linear classification on the checkpoint is trained for 1 epoch. The results are shown in Figure~\ref{fig:cutmix_example_ablation}. Cut-mixed examples in various settings improves performance. We used  $\Beta(5, 3)$ in our experiments.

\section{Experiments details for transfer learning}

\paragraph{Linear classification on ImageNet}
The main network is fixed, and global average pooling features (2048-D) of ResNet-50 are extracted. We train for 100 epochs. For \shortname{}$_{\text{MoCo}}$, we follow the block-decay training schedule of~\cite{he2020momentum,chen2020improved} with an initial
learning rate of 30.0 and step decay with a factor of 0.1 at $[60, 80]$.
For  \shortname{}$_{\text{DCluster}}$, we follow the cosine-decay training schedule of~\cite{caron2020unsupervised} with an initial learning rate of 0.3. The logistic regression classifier is trained using SGD with a momentum of 0.9.

\paragraph{Linear classification  on VOC07}
For training linear SVMs on VOC07, we follow the procedure in~\cite{goyal2019scaling,li2020prototypical} and use the LIBLINEAR package~\cite{fan2008liblinear}. We pre-process all images by resizing to 256 pixels along the shorter side and taking a $224 \times 224$ center crop. The linear SVMs are trained on the global average pooling features of ResNet-50.

\paragraph{Linear classification on Cifar10 and Cifar100}
We trained a linear classifier on features extracted from the frozen pre-trained network. We used Adamax to optimize the softmax cross-entropy objective for 20 epochs, a batch size of 256, a learning rate [0.1,0.01,0.001] and decay at [7, 14] with a factor of 0.1. All images were resized to $224$ pixels (after which we took a $224 \times 224$ center crop), and we did not apply data augmentation.


\paragraph{Semi-supervised learning on ImageNet} We follow~\cite{caron2020unsupervised} to finetune ResNet-50 with pretrained weights on a subset of ImageNet with labels. We optimize the model with SGD, using a batch size of 256, a momentum of 0.9, and a weight decay of 0.0005. We apply different learning rate to the ConvNet and the linear classifier. The learning rate for the ConvNet is 0.01, and the learning rate for the classifier is 0.1 (for 10\% labels) or 1 (for 1\% labels). We train for 20 epochs, and drop the learning rate by 0.2 at 12 and 16 epochs.

\paragraph{Object detection on VOC} We follow~\cite{chen2020improved} to use the R50-FPN backbone for the Faster R-CNN detector available in the Detectron2 codebase~\cite{wu2019detectron2}. We freeze all the conv layers and also fix the BatchNorm parameters. The model is optimized with SGD, using a batch size of 8, a momentum of 0.9, and a weight decay of 0.0001. The initial learning rate is set as 0.05. We finetune the models for 15 epochs, and drop the learning rate by 0.1 at 12 epochs.

\section{Experiments on Fine-tuning}

We fine-tuned the entire network using the weights of the pre-trained network as
initialization. We trained for 20 epochs at a batch size of 256 using Adamax, decayed at [7,14] with a factor of 0.1. We grid search learning rate over [0.0005, 0.001,  0.01]. The results are shown in Table~\ref{tab:finetune_cls}. Our \shortname{} consistently improves their original counterparts for both datasets.

\begin{table}[t!]
\small
\centering
\begin{tabular}{ @{\hspace{-0pt}}l@{\hspace{5pt}}c@{\hspace{5pt}}@{\hspace{10pt}}c@{\hspace{10pt}}c@{\hspace{10pt}}}
\toprule
 Method   & Epoch & C10 &  C100  \\ 
\hline
Supervised~\cite{chen2020simple} & - & 97.5 & 86.4 \\
Supervised$^\dagger$ & - & 96.7 & 83.5 \\
Random Init~\cite{chen2020simple} & - & 95.9 & 80.2 \\
\hline
SimCLR~\cite{chen2020simple} & 1000  & 97.7 & 85.9 \\
SwAV \!(B=256)$^\dagger$~\cite{caron2020unsupervised} & 200  & 96.6 & 82.7 \\
SwAV \!(B=4096)$^\dagger$~\cite{caron2020unsupervised} & 200  & 96.4 & 83.2 \\
\hline
 MoCo-v2~~ &  200  &  95.6 & 80.8\\
\rowcolor{Gray}
\cellcolor{white}
   \shortname{}$_{\text{MoCo}}$  &  200  & \textcolor{blue}{96.1} &  \textcolor{blue}{81.3}
  \\ \hline
 MoCo-v2~~ &  800 & 96.1 & 83.0 \\
\rowcolor{Gray}
\cellcolor{white}
   \shortname{}$_{\text{MoCo}}$  &  800   & \textcolor{blue}{96.5}  & \textcolor{blue}{83.5} 
  \\ \hline
 DeepCluster-v2~~ &  200 & 96.2 & 82.1\\
\rowcolor{Gray}
\cellcolor{white}
   \shortname{}$_{\text{DCluster}}$  &  200  &    \textcolor{blue}{97.2} &  \textcolor{blue}{84.4}  \\  
\rowcolor{Gray}
\cellcolor{white}
   \shortname{}$_{\text{DCluster}}$(8-crop \!)  &  200    &   \textcolor{blue}{ 97.0} &  \textcolor{blue}{85.5}    \\  
\rowcolor{Gray}
\cellcolor{white}
   \shortname{}$_{\text{DCluster}}^{+}$(8-crop \!)   &  200   &   \textcolor{blue}{96.9}  &  \textcolor{blue}{84.9}   \\    
\bottomrule
\end{tabular}

\vspace{-1mm}
\caption{Image classification performance on fine-tuning the entire ResNet-50 network. All numbers for baselines are from~\cite{chen2020simple} , except that we use the released pretrained model for SwAV. $^\dagger$ indicates the results based on our runs using the same training schedules.}
\label{tab:finetune_cls}
\vspace{-0mm}
\end{table}

\end{document}